\documentclass[10pt,twocolumn,letterpaper]{article}

\usepackage{iccv}
\usepackage{times}
\usepackage{epsfig}
\usepackage{graphicx}
\usepackage{amsmath}
\usepackage{amssymb}
\usepackage{algorithm2e}
\usepackage{colortbl}

\usepackage[normalem]{ulem}

\graphicspath{{figures/}}

\usepackage{algorithm}

\usepackage[pagebackref=false,breaklinks=true,letterpaper=true,colorlinks,bookmarks=false]{hyperref}

\iccvfinalcopy 

\def\iccvPaperID{851} 

\def\etal{{et al}\onedot}

\def\rr#1{{\bf \color{red} #1}}
\def\br#1{{\bf \color{blue!80} #1}}
\def\gr#1{{\bf \color{black!10!green!90} #1}}
\newlength{\oldtextfloatsep}

\ificcvfinal\fi

\begin{document}
\newcommand{\twofigures}[3]{
            \centerline{{\includegraphics[width=#3]{#1}}~~{\includegraphics[width=#3]{#2}}}}

\newcommand{\threefiguresh}[4]{
            \centerline{{\includegraphics[height=#4]{#1}}{\includegraphics[height=#4]{#2}}{\includegraphics[height=#4]{#3}}}}

\newcommand{\threefigures}[4]{
            \centerline{{\includegraphics[width=#4]{#1}}~~{\includegraphics[width=#4]{#2}}~~{\includegraphics[width=#4]{#3}}}}

\newcommand{\threefigureslbl}[5]{
            \centerline{(#5){\includegraphics[width=#4]{#1}}~~{\includegraphics[width=#4]{#2}}~~{\includegraphics[width=#4]{#3}}}}

\newcommand{\lthreefigures}[4]{
            \centerline{{\includegraphics[width=#4]{#1}}{\includegraphics[width=#4]{#2}}{\includegraphics[width=#4]{#3}}
			\makebox[0.33\columnwidth][c]{(a)}\makebox[0.33\columnwidth][c]{(b)}\makebox[0.33\columnwidth][c]{(c)}}}

\newcommand{\fourfigures}[5]{
            \centerline{{\includegraphics[width=#5]{#1}}~~{\includegraphics[width=#5]{#2}}~~{\includegraphics[width=#5]{#3}}~~{\includegraphics[width=#5]{#4}}}}

\newcommand{\comment}[1]{}

\title{Online Object Tracking with Proposal Selection}

\author{Yang Hua ~~~~~~Karteek Alahari ~~~~~~Cordelia Schmid\\
~\\
Inria\thanks{LEAR team, Inria Grenoble Rh\^one-Alpes, Laboratoire Jean Kuntzmann, CNRS, Univ. Grenoble Alpes, France.}
}


%

\maketitle
\thispagestyle{empty}

\begin{abstract}
\vspace{-0.3cm}
Tracking-by-detection approaches are some of the most successful object
trackers in recent years. Their success is largely determined by the detector
model they learn initially and then update over time. However, under
challenging conditions where an object can undergo transformations, e.g.,
severe rotation, these methods are found to be lacking. In this paper, we
address this problem by formulating it as a proposal selection task and making
two contributions. The first one is introducing novel proposals estimated from
the geometric transformations undergone by the object, and building a rich
candidate set for predicting the object location. The second one is devising a
novel selection strategy using multiple cues, i.e., detection score and
edgeness score computed from state-of-the-art object edges and motion
boundaries. We extensively evaluate our approach on the visual object tracking
2014 challenge and online tracking benchmark datasets, and show the best
performance.
\end{abstract}

\vspace{-0.3cm}
\section{Introduction}
\label{sec:intro}
\vspace{-0.2cm}
Over the past few years, the tracking-by-detection framework has emerged as one
of the successful paradigms for tracking
objects~\cite{Kristan14,Pang13,Song13,WuY13}. It formulates the tracking
problem as the task of detecting an object, which is likely to undergo changes
in appearance, size, or become occluded, over
time~\cite{Avidan07,Babenko11,Godec11,Grabner08,Hare11,Kalal12,Leibe08,Ramanan07,Supancic13,Wu07}.
These approaches begin by training an object detector with an initialization
(in the form of a bounding box) in the first frame, and then update this model
over time. Naturally, the choice of exemplars used to update and improve the
object model is critical~\cite{Hua14,Supancic13,Zhang14}.

\begin{figure}[t]
\begin{center}
\twofigures{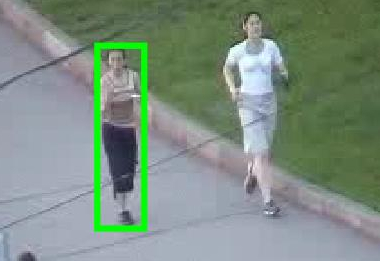}{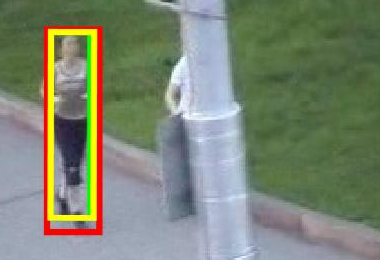}{0.5\columnwidth}
\vspace{0.1cm}
\twofigures{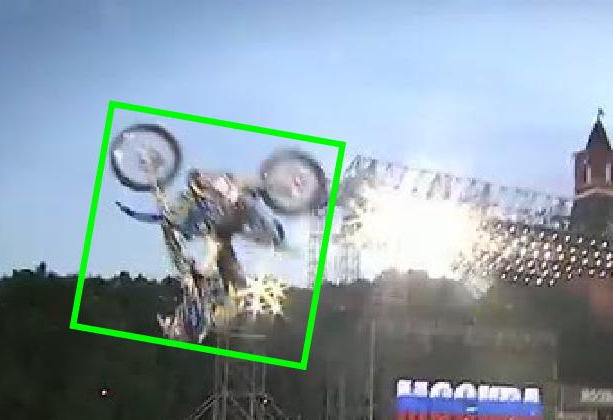}{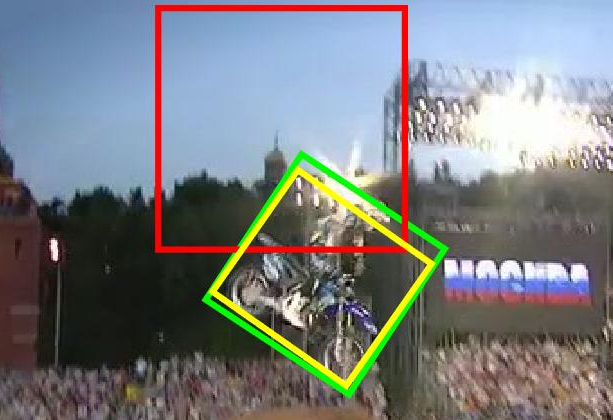}{0.5\columnwidth}
\end{center}
\vspace{-0.8cm}
\caption{Sample frames (cropped) from the {\em jogging} (top row) and {\em
motocross} (bottom row) sequences~\cite{Kristan14}. The ground truth
annotation (green) in the first frame (left) is used to train our tracker and the
winner~\cite{Danelljan14} of VOT2014 challenge~\cite{Kristan14}. We
show these two tracking results (right) on another frame in the sequence. Our
method (yellow) successfully tracks objects undergoing deformations
unlike~\cite{Danelljan14} (red). {\it Best viewed in pdf.} \vspace{-0.55cm}}
\label{fig:defm}
\end{figure}

Consider the {\em motocross} example shown in Figure~\ref{fig:defm}. Here, the
target (biker and his motorbike) undergoes several deformations as the biker
performs an acrobatics routine, which leads to significant changes in the
aspect ratio as well as the rotation angle of the bounding box.
State-of-the-art tracking methods, e.g.,~\cite{Hua14,Supancic13}, rely on
axis-aligned bounding boxes and thus are ill-equipped to capture the accurate
extents of the object in such scenarios. In this paper, we aim to address this
issue by treating the tracking problem as the task of selecting the best
proposal containing the object of interest from several candidates. To this
end, we propose a novel method to generate candidates which capture the extents
of the object accurately, by estimating the transformation(s) undergone by the
object.

The focus of this paper is the problem of tracking a single target in monocular
video sequences. The target is provided as a ground truth annotation in the
first frame, as in a standard tracking-by-detection
setup~\cite{Hare11,Kalal12}. We learn an object detector model from this
annotation and evaluate it in subsequent frames to propose candidate locations
that are likely to contain the target. While these proposals are sufficient to
track objects undergoing a small subset of transformations over time, such as
translation shown in Figure~\ref{fig:defm}-top, they cannot handle generic
transformations, e.g., similarity transformation shown in
Figure~\ref{fig:defm}-bottom. We address this problem by: (i) introducing novel
additional proposals to improve the candidate location set, and (ii) using
multiple cues to select the proposal that is most likely to contain the target.
Additional proposals are computed by robustly estimating the parameters of
similarity transformation (i.e., scaling, translation, rotation)
that the target is likely to have undergone, with a Hough transform voting
scheme on the optical flow. With these parameters, we propose several candidate
locations for the target. Thus, for every frame of the video sequence, our
candidate set consists of object detector boxes as well as geometry estimation
proposals. Note that state-of-the-art tracking-by-detection approaches are
limited to detector proposals alone. The second contribution we make is in
selecting the best proposal from the candidate set using multiple cues --
objectness scores~\cite{Zitnick14} computed with edge responses~\cite{Dollar13}
and motion boundaries~\cite{Weinzaepfel15}, and the detection score.

We evaluate the performance of our approach exhaustively on two recent
benchmark datasets, namely the visual object tracking (VOT) 2014 challenge
dataset~\cite{Kristan14} and the online tracking benchmark (OTB)~\cite{WuY13}.
Furthermore, we compare with all the 38 trackers evaluated as part of the VOT2014 challenge, and with the best performers~\cite{Hare11,Jia12,Zhong12} among
the 29 methods analyzed in the OTB comparison paper~\cite{WuY13}. Our method
shows the top performance on both these challenging datasets, and is in
particular 29.5\% better than the current leader~\cite{Danelljan14} of the
VOT2014 challenge.

The remainder of the paper is organized as follows. In
Section~\ref{sec:related} we discuss related work. The details of our novel
proposals and the multiple cues to select the best candidate are explained in
Section~\ref{sec:hough}. Section~\ref{sec:implement} discusses the
implementation details of the methods. In Section~\ref{sec:expts} we present an
extensive evaluation of the proposed method and compare it with the state of
the art. This section also discusses the datasets and evaluation protocols.
Concluding remarks are made in Section~\ref{sec:summary}.

\vspace{-0.2cm}
\section{Related Work}
\label{sec:related}
\vspace{-0.2cm}
Two of the key elements of any tracking algorithm are, how the object of
interest is represented, and how this representation is used to localize it
in each frame. Several methods have been proposed on these two fronts. Object
representation has evolved from classical colour histograms~\cite{Comaniciu03}
to models learned from generative~\cite{Lee05,MeiX11,Ross08} or
discriminative~\cite{Avidan07,Collins05,Hare11,Hua14} approaches. In the
context of localizing the object given a representation, methods such
as~\cite{Badrinarayanan07,Birchfield98,Isard98,ParkDW12,Perez04,Spengler03,Stenger09}
have incorporated multiple cues and fused their results with Markov chain Monte
Carlo~\cite{ParkDW12} or particle filtering~\cite{Isard98}. More recently,
inspired by the success of object detection
algorithms~\cite{Everingham10,Felzenszwalb10}, the tracking-by-detection
approach~\cite{Avidan07,Babenko11,Hare11,Kalal12,Leibe08,Ramanan07,Supancic13,Zhang14}
has gained popularity. In fact, methods based on this paradigm are ranked
among the top performers on evaluation benchmarks~\cite{Pang13,WuY13}.

Tracking-by-detection methods learn an initial discriminative model of the
object from the first frame in the sequence (e.g., with a support vector
machine (SVM)~\cite{Hare11,Supancic13}), evaluate it to detect the most likely
object location in subsequent frames, and then update the object model with
these new detections. {\it Struck}~\cite{Hare11} is an interesting variant of
this general framework, using a structured output formulation to learn and
update the detector. This shows state-of-the-art results in several scenarios, but
lacks the capability of handling severe occlusions and suffers from drift, as
discussed in~\cite{Hua14}.

To avoid the well-known problem of drift~\cite{Matthews04}, some methods
explicitly detect tracking failures and occlusions with
heuristics~\cite{Zhong12}, combine detector- and tracker-based
components~\cite{Kalal12}, control the training set with
learning~\cite{Supancic13}, or maintain a pool of
trackers~\cite{Hua14,Kwon11,Zhang14}. Kalal~\etal~\cite{Kalal12} combine
predictions from a variant of the Lukas-Kanade tracker and an incrementally updated
detector. More specifically, the results of the tracker provide training data
to update the detector model, and the detector re-initializes the tracker when
it fails. This interleaving of tracker and detector, although interesting, is
restrictive as the object model is a set of fixed-size template patches, and
thus, not robust to severe changes in object size.

Supancic and Ramanan~\cite{Supancic13} propose a self-paced learning scheme to
select training examples for updating the detector model. A pre-fixed number of
(top~$k$) frames are chosen carefully according to an SVM objective function to
avoid corrupting the training set. In other words, only frames with confident
detections are used to update the model. This approach, however, lacks the
ability to handle commonly-occurring scenarios where the object is undergoing a
geometric transformation, such as the rotation shown in
Figure~\ref{fig:defm}-bottom.  Hua~\etal~\cite{Hua14} address this by using
long-term trajectories to estimate the transformations undergone by the object.
However, their approach is: (i) inherently offline, requiring all the frames in
the video sequence to track, (ii) not robust to significant changes in scale,
(iii)~maintains a pool of trackers, all of which need to be evaluated
independently in every frame.

Despite the general success of tracking-by-detection approaches, they have an
important shortcoming. They all rely solely on a detector, and are thus unable
to cope with transformations an object can undergo. On the other hand, key point
based trackers like CMT~\cite{Nebehay14} detect key points in a frame, match
them to the next frame and can estimate transformations, such as changes in scale and rotation. However,
they are highly sensitive to the success of key point detection algorithms and
lack a learned model. Indeed, this is evident from the poor performance
on the VOT dataset (see Section~\ref{sec:results}). The main focus of this
paper is to address the gap between these two paradigms. Inspired by successful
object proposal methods for segmentation and
detection~\cite{Alexe12,Carreira10}, we pose the tracking problem as the task
of finding the best proposal from a candidate set. In contrast to previous
work, we introduce novel proposals estimated from the geometric transformations
undergone by the object of interest, and build a rich candidate set for
predicting the object location, see Figure~\ref{fig:proposals}. In essence, our
candidate set consists of proposals from the detector as well as the geometry
estimation model. We then devise a novel scoring mechanism using standard
object detector score, together with object edgeness~\cite{Zitnick14} computed from
state-of-the-art edge detector SED~\cite{Dollar13} and motion boundary
detector~\cite{Weinzaepfel15} to find the most likely location of the object.

\vspace{-0.2cm}
\section{Proposal Selection for Tracking}
\label{sec:hough}
\vspace{-0.2cm}
The main components of our framework for online object tracking are: (i)
learning the initial detector, (ii) building a rich candidate set of object
locations in each frame, consisting of proposals from the detector as well as
the estimated geometric transformations, (iii) evaluating all the proposals in
each frame with multiple cues to select the best one, and (iv) updating the detector model. We now
present all these components and then provide implementation details in
Section~\ref{sec:implement}.

\vspace{-0.15cm}
\subsection{Initial detector}
\label{sec:initial}
\vspace{-0.15cm}
We learn the initial detector model with a training set consisting of one
positive sample, available as a bounding box annotation in the first frame, and
several negative bounding box samples which are automatically extracted from
the entire image. We use HOG features~\cite{Dalal05,Felzenszwalb10} computed for these
bounding boxes and learn the detector with a linear SVM, similar to other
tracking-by-detection approaches~\cite{Hua14,Supancic13}. The detector is then
evaluated on subsequent frames to estimate the candidate locations of the
object. Rather than make a suboptimal decision by choosing the best detection
as the object location in a frame, we extract the top~$k$ detections and build
a candidate set. We augment this set with proposals estimated from
the transformations undergone by the object, as described in the following
section.

\begin{figure}[t]
\begin{center}
\twofigures{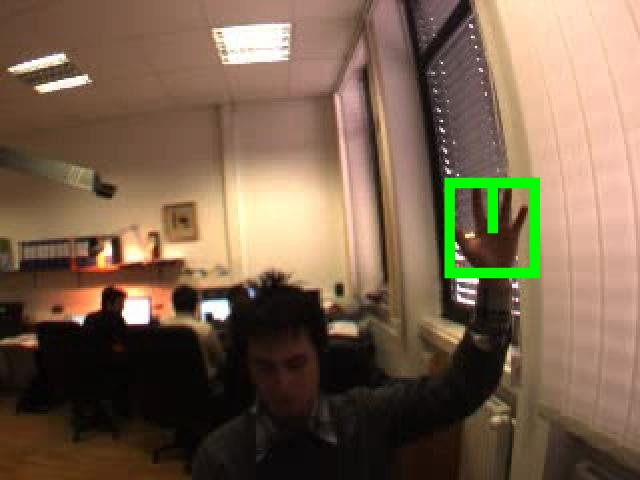}{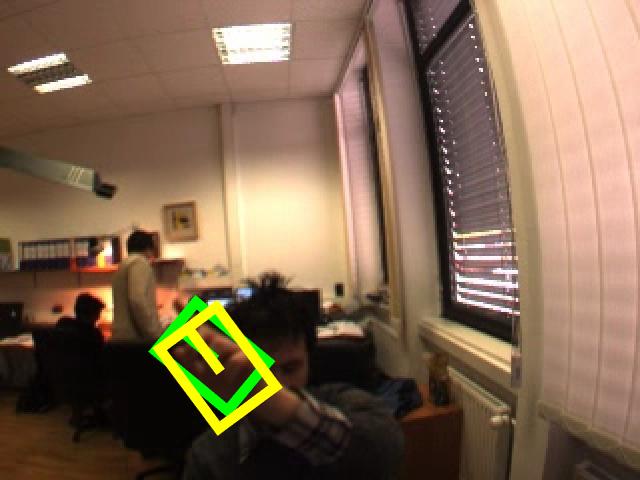}{0.5\columnwidth}
\twofigures{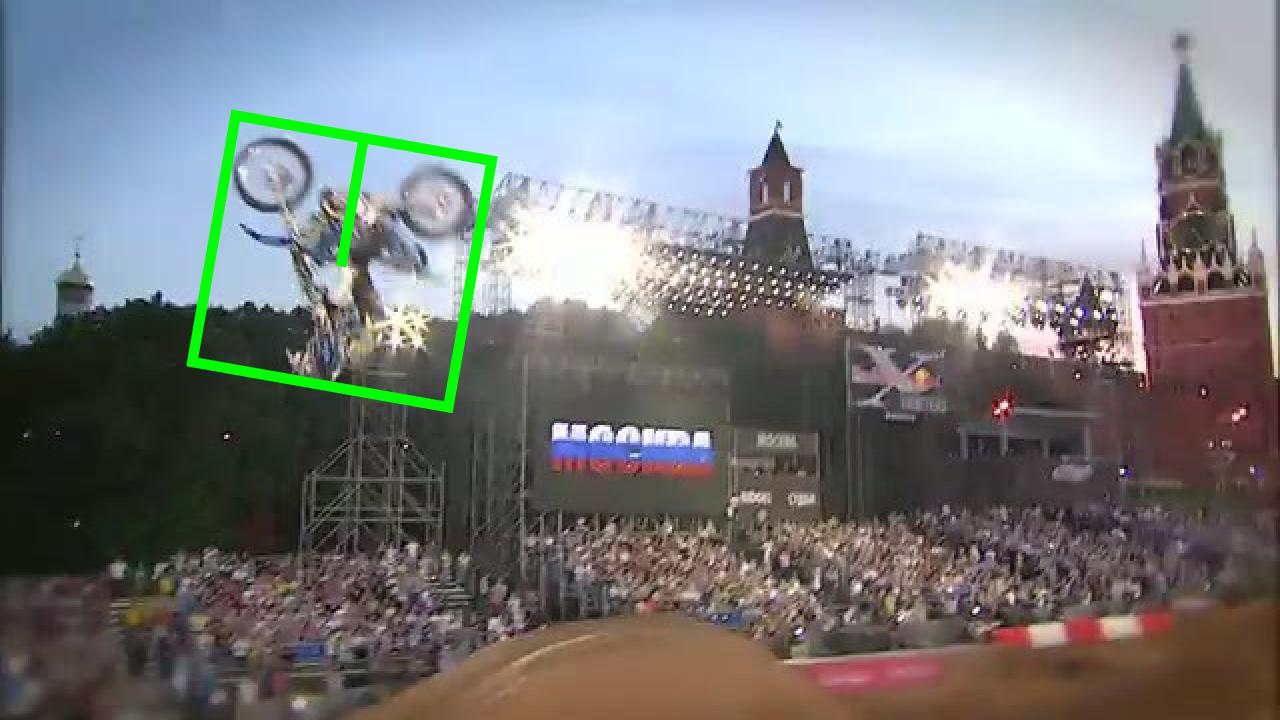}{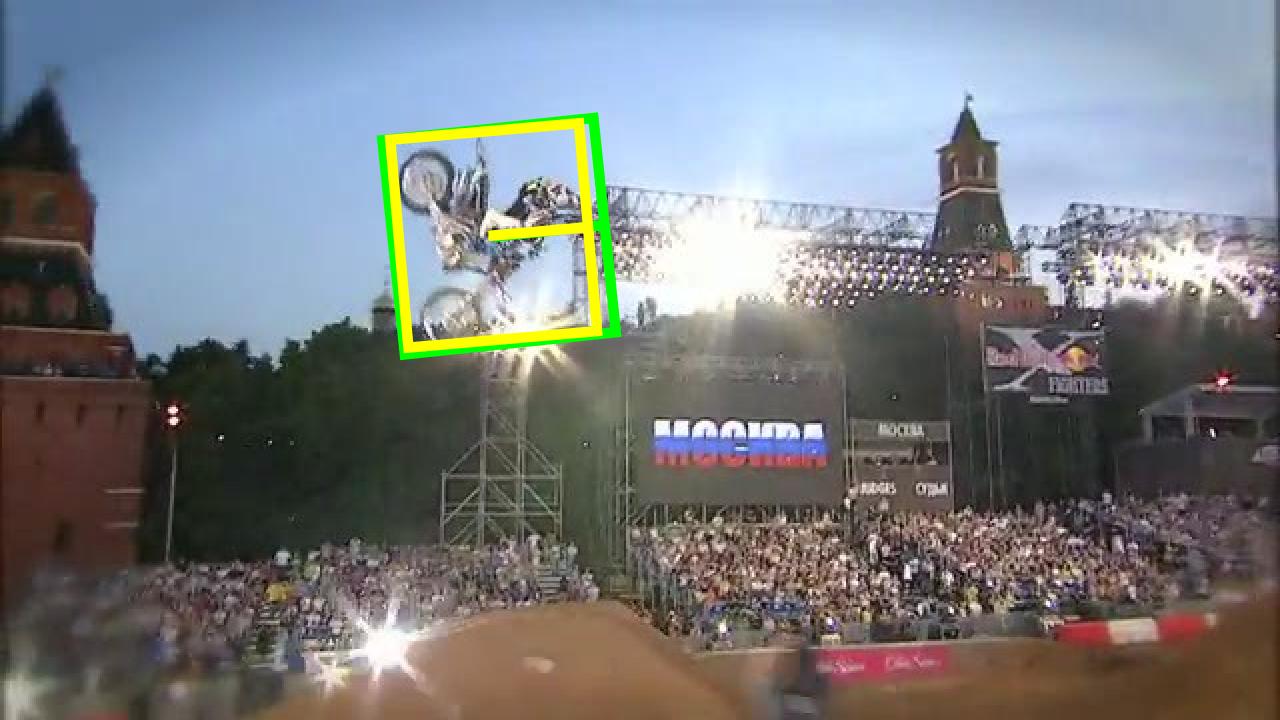}{0.5\columnwidth}
\end{center}
\vspace{-0.8cm}
\caption{Two sequences from the VOT dataset~\cite{Kristan14}. In each row, we
show the ground truth (in green) in the first frame (left) and the best geometry proposal
in a subsequent frame (in yellow in the right image). We show a vertical line on
each box to visualize the rotation angle with respect to image edges. \vspace{-0.4cm}}
\label{fig:proposals}
\end{figure}

\vspace{-0.1cm}
\subsection{Estimating geometry and proposals}
\label{sec:estimate}
\vspace{-0.15cm}
In the examples shown in Figure~\ref{fig:proposals}, the object of interest is
undergoing a geometric transformation (rotation in these examples). None of
the top detector proposals can capture the object accurately in this case. To
address this issue, we explicitly estimate the transformation undergone by the
object and then use it to enrich the candidate set with additional geometric
proposals.

We represent the geometric transformation with a similarity
matrix~\cite{Hartley04}. Note that other transformations like homography can
also be used. The similarity transformation is defined by four parameters --
one each for rotation and scale, and two for translation. In this work, we
estimate them with a Hough transform voting scheme using frame-to-frame optical
flow correspondences. We begin by computing the optical flow between frames
$t-1$ and $t$ with a state-of-the-art flow estimation method~\cite{Brox11}. The
pixels within the object bounding box with flow values in frame $t-1$ then give
us corresponding pixels in frame $t$. With a pair of these matches, given by
two pixels in $t-1$ which are sufficiently distant from each other, we estimate
the four scalar parameters in the similarity matrix~\cite{Hartley04}. Every
choice of a pair of corresponding matches gives us an estimate of the four
parameters. We then use the Hough transform~\cite{Hough62}, wherein each pair
of point-matches votes for a (4D) parameter set, to find the top~$k$ consistent
parameter sets. This voting scheme allows us to filter out the incorrect
correspondences due to common errors in optical flow estimation.

To sum up, we estimate the $k$ most likely geometric transformations undergone
by the object as the parameters with the top~$k$ votes. Each one of these
transformations results in a candidate location of the object in $t$, by
applying the corresponding similarity matrix to the object bounding box in
$t-1$. Before adding all these $k$ proposals to the candidate set, we perform an additional filtering
step to ensure further robustness. To this end, we discard all the proposals:
(i) which have a low confidence, given by the number of votes, normalized by the
total number of point-matches, and (ii) those where the estimated angle is
significantly different from estimations in earlier frames.  We determine both
these threshold values with Kalman filtering~\cite{Kalman60}.  A threshold
$\tau_t$, used for filtering proposals in frame $t+1$, is given by:
\vspace{-0.1cm}
\begin{eqnarray}
\tau_t &=& \alpha \tau_{t-1} + k_t (d_t - \alpha \beta \tau_{t-1}),\\
k_t &=& (p_{t-1} + q) / (p_{t-1} + q + r),\\
p_t &=& (1-k_t) (p_{t-1} + q),
\vspace{-0.2cm}
\end{eqnarray}
where $\alpha$, $\beta$, $q$, $r$ are scalar parameters set empirically (see
Section~\ref{sec:implement}), $k_t$ is the Kalman gain, $p_t$ is the error
estimate. The filter is initialized with $\tau_0 = 0$, $p_0 = 0.1$. Here, $d_t$
is either the confidence score or the estimated angle of the selected proposal in frame $t$
to determine the respective threshold.


\begin{figure}[t]
\begin{center}
\includegraphics[width=\columnwidth]{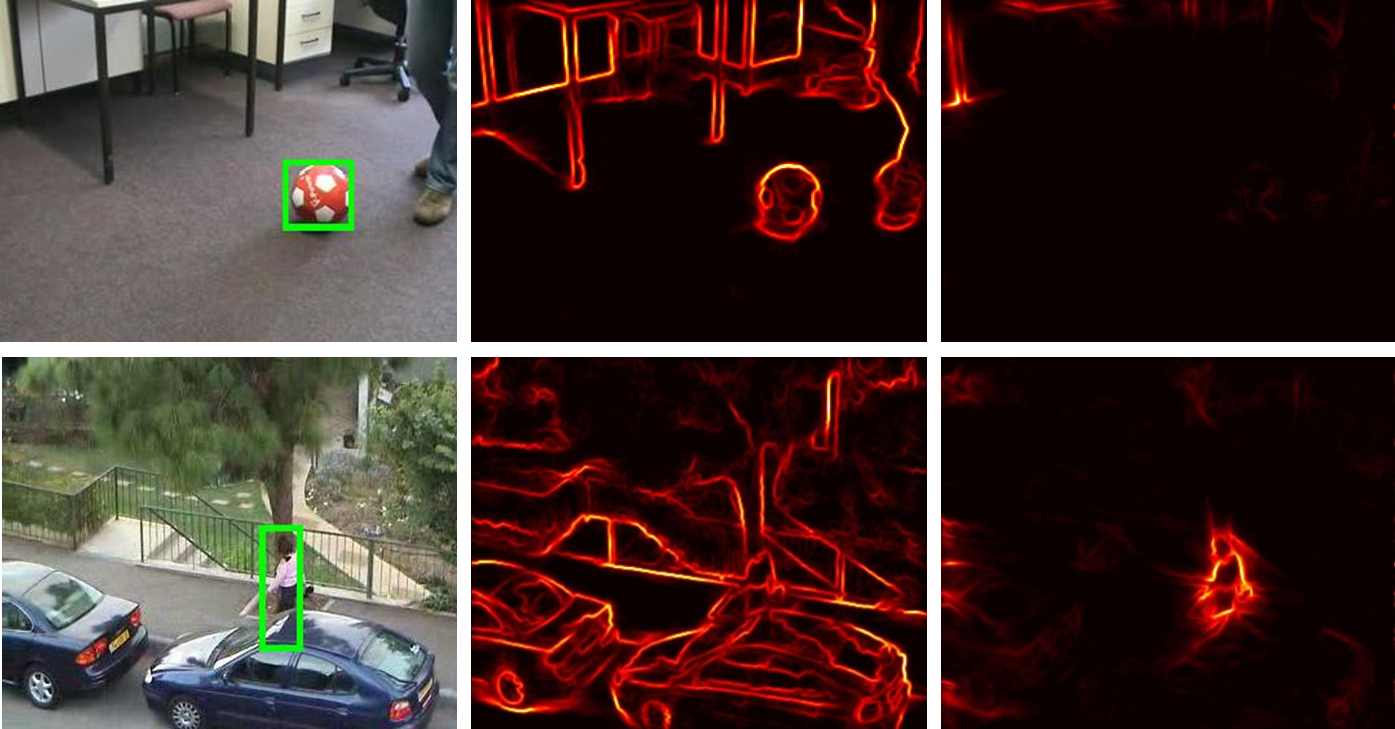}
\end{center}
\vspace{-0.3cm}
\caption{Two examples of edge and motion boundary responses.
In each row, from left to right, we show the original image (with the object in a green box),  edges and motion boundaries. The edges provide a stronger response in
the example in the first row while motion boundaries are more effective in the
example in the second row. \vspace{-0.5cm}}
\label{fig:edgemotion}
\end{figure}


\subsection{Selecting proposals}
\label{sec:evaluate}
\vspace{-0.2cm}
At the end of the geometry estimation step, we have $k$ proposals from the
detector and (at most) $k$ proposals from the similarity transformation in
frame $t$. Now, the task is to find the {\it best} proposal most likely to
contain the object in this frame. We use three cues, detection
confidence score, objectness measures computed with object edges and motion
boundaries, for this task. We first use the (normalized) detection
confidence score computed for each proposal box with the SVM learned from the
object annotation in the first frame and updated during tracking. This provides
information directly relevant to the object of interest in a given sequence.

In cases where the detection scores of all the candidate boxes are
statistically similar, we use cues extracted from object edges~\cite{Dollar13}
and motion boundaries~\cite{Weinzaepfel15} in the image, referred to as
edgeness scores. In other words, when the detection scores are
inconclusive to choose the best proposal, we rely on edgeness measure. Here,
all the top candidates contain the object of interest to some extent, and the
task is to choose the one that best contains the object---a scenario
well-suited for edgeness cue because a proposal box which accurately
localizes the entire object has a stronger edge response compared to a box
which overlaps partially with the object, thus resulting in a weaker edge
score. 
The edgeness score is given by the number of edges (or motion boundaries) within
a proposal box after discarding contours that intersect with the box's
boundary, as in~\cite{Zitnick14}. We compute edgeness scores with edges and
motion boundaries separately and use the one with a stronger response. As shown
in the examples in Figure~\ref{fig:edgemotion}, these two cues are
complementary. Then, the proposal with the best edgeness score is
selected. 

We follow this approach, rather than using a combined score, e.g., a weighted
combination of SVM and edgeness scores, because setting this weight manually is
suboptimal, and learning it is not possible due to the lack of standard
training-validation-test datasets for tracking. In contrast, our approach uses
object-specific SVM and generic edgeness scores effectively, and circumvents
the need for a manually set weight to combine scores.

\subsection{Updating the detector}
\vspace{-0.1cm}
\label{sec:update}
Having computed the best proposal containing the object, box$_t$, we use it as
a positive exemplar to learn a new object model. In practice, we update the
current model incrementally with this new sample; see
Section~\ref{sec:implement}. This new detector is then evaluated in frame
$t+1$, to continue tracking the object. Algorithm~\ref{algo:method} summarizes
our overall approach.

\vspace{-0.1cm}
\section{Implementation Details}
\label{sec:implement}
\vspace{-0.2cm}
\paragraph{Detector.}
The initial detector is trained with the one positive sample (provided as
ground truth box in the first frame) and several negative examples extracted
from the frame. We harvest bounding boxes that overlap less than 50\% with the
ground truth for the negative set. The regularization parameter in the SVM cost
function is set to 0.1 in all our experiments. This cost function is minimized
with LIBLINEAR~\cite{Fan08}. We perform three rounds of hard negative mining to
refine the detector. The detector scores are calibrated with Platt's
method~\cite{Platt99} using jittered versions of the positive sample (100
positive and negative samples each). For the VOT dataset, an additional step is
introduced to train with rotated bounding box annotations. We estimated the
rotation angle between one edge of the box and the corresponding image
axis,\footnote{Note that this estimation was done only in the first frame or
whenever the tracker is restarted after a failure, i.e., when we have access to
the ground truth box.} rectified the image with this angle, and extracted an
axis-aligned box as the positive sample.

The detector was evaluated at seven scales: \{0.980, 0.990, 0.995, 1.000, 1.005, 1.010, 1.020\}, in every frame. 
This is done densely in each scale, with a step size of 2 pixels.
In the rotated bounding box case, we
rectify the image with the angle estimated during training, before evaluating
the detector. We set $k=5$ to select the top 5 detection results and also (at
most) 5 geometry proposals to form the candidate set. The best proposal box$_t$
selected from all the candidates in a frame is used to update the detector with
a warm-start scheme~\cite{Fan08,Supancic13}. We then perform hard negative
mining and calibrate the scores to complete the detector update step. For
computational efficiency, the detector is evaluated in a region around the
previously estimated object location, instead of the entire image.

\vspace{-0.3cm}
\paragraph{Hough voting.} The geometric transformation is estimated only when
the mean $\ell_2$ norm of optical flow in box$_{t-1}$ is larger than 0.5. This
ensures that there is sufficient motion in the frame to justify estimating a
transformation. 
We use two pixels in $t-1$ and their corresponding
points in $t$ to vote for the geometric transformation parameters. In practice,
pixels that are at least 25 pixels apart from each other in $t-1$ were chosen
randomly for a reliable estimation. To aggregate the votes, a pseudo-random
hash function of bin values is used to vote into a one-dimensional hash table,
instead of a four-dimensional array, similar to~\cite{Lowe04}. The bin size for
scale is set to 0.1, and 2.0 for angle and the two translation parameters.
Finally, we take the least-squares solution of all the candidates that vote for
a particular bin, making our geometry estimation further robust.

\vspace{-0.3cm}
\paragraph{Pruning proposals.} The parameters of the Kalman filter to discard
weak geometry proposals are set as: $\alpha = 1$, $\beta = 1$, process error
variance $q = 0.001$, and measurement error variance $r = 0.01$ in all our
experiments. To find the best proposal from a candidate set, we propose two
cues: detection and edgeness scores. Edgeness scores are used to determine the
best proposal when the detection scores are inconclusive, i.e., when one or
more proposals differ from the best detection score by at most 1\%. However,
edgeness scores can often be corrupted by noise, e.g., low-quality images that
are common in benchmark datasets. We ensure that these scores are used only
when they are comparable to the mean score of the previous 5 frames, i.e., the
difference between the highest edgeness score and the mean is less than twice
the variance. This filtering step is performed separately for edgeness computed
with object edges and motion boundaries.

\setlength{\textfloatsep}{5pt}
\begin{algorithm}[t]
\KwData{Image frames $1 \ldots n$, Ground truth box$_1$ in frame 1}
\KwResult{Object location box$_t$ in frames $t = 2\ldots n$}
 Learn initial detector model in frame $1$ (\S\ref{sec:initial})\\
 \For {$t = 2 \ldots n$}{
  candidate set $\mathcal{C}_t \gets $ Top-$k$ detections in frame $t$ (\S\ref{sec:initial})\\
  $\mathcal{H}_t \gets $ Geometry proposals in frame $t$ (\S\ref{sec:estimate})\\
  $\mathcal{C}_t \gets \mathcal{C}_t \cup \mathcal{H}_t$\\
  box$_t \gets $ Best proposal in $\mathcal{C}_t$ (\S\ref{sec:evaluate})\\
  Update detector model with box$_t$ (\S\ref{sec:update})
 }
\caption{: Our approach for online object tracking.}
\label{algo:method}
\vspace{-0.15cm}
\end{algorithm}

\vspace{-0.2cm}
\section{Experiments}
\label{sec:expts}
\vspace{-0.2cm}
We now present our empirical evaluation on two state-of-the-art benchmark
datasets and compare with several recent methods.  The results of all the
methods we compare with are directly taken from the respective toolkits. Additional results and videos are provided on the project website~\cite{project}.

\subsection{Datasets and Evaluation}
\label{sec:dataset}
\vspace{-0.2cm}
Much attention has been given to developing benchmark datasets and measures to
evaluate tracking methods in the past couple of years~\cite{Smeulders14}. For
example, Wu~\etal~\cite{WuY13} and Song and Xiao~\cite{Song13} introduced
benchmark datasets and evaluated several state-of-the-art tracking algorithms.
Kristan~\etal~\cite{Kristan14} have organized visual object tracking (VOT)
challenges in 2013 and 2014, where several challenging sequences were used to
evaluate methods. We have evaluated our method on the VOT2014~\cite{Kristan15}
and the online tracker benchmark (OTB)~\cite{WuY13} datasets, following their
original protocols.

\setlength{\textfloatsep}{\oldtextfloatsep}
\vspace{-0.4cm}
\paragraph{VOT2014 dataset.}
As part of the visual object tracking challenge held at ECCV
2014~\cite{Kristan14}, the organizers released a set of 25 challenging video
sequences. They were chosen from a candidate set of 394 sequences, composed of
examples used in several previous works such as VOT2013 benchmark dataset,
ALOV~\cite{Smeulders14}, OTB~\cite{WuY13}, as well as a few unpublished ones.
This choice was made by: (i) discarding sequences shorter than 200 frames, and
(ii) ensuring that the selected sequences contain well-defined targets and
represent challenging scenarios, e.g., clutter, object deformation, change in
aspect ratio. The frames in all the sequences are
manually annotated. In a departure from annotations in other datasets, where
the bounding box enclosing the target is axis-aligned, this benchmark uses
rotated boxes in order to handle targets that are deforming, rotating or
elongated, as shown in Figure \ref{fig:proposals}. This new annotation makes the dataset a
more challenging setting to evaluate trackers. In addition to this, all the
frames are labelled with visual attributes: occlusion, illumination change,
object motion, object size change, camera motion and neutral.

We follow the protocol described in~\cite{Kristan15} to evaluate our tracking
algorithm, and compare it with several state-of-the-art trackers. This
evaluation scheme uses accuracy and robustness measures to compare trackers.
Accuracy is computed as the mean intersection over union score with the ground
truth bounding box over the entire sequence (while discarding a few frames
immediately following a tracking failure), and robustness is the number of
times the tracker has failed. A tracking failure is signalled in a frame $s$ if
the predicted box does not overlap with the ground truth annotation. In this
case, the tracker is restarted from scratch with the ground truth annotation in
frame $s+5$.

The scores are averaged over several runs of the tracker to account for any
stochastic behaviour. The overall rank of a tracker is determined by first
ranking trackers on accuracy and robustness separately on each attribute set,
then averaging the rank over the attributes, and then taking the mean rank over
the two performance measures. If a set of trackers show similar performance
(which is measured with statistical significance and practical difference
tests), their ranks are averaged. This rank correction is performed separately
for accuracy and robustness, and the overall average rank is recomputed as
mentioned above. We used the
toolkit\footnote{{http://www.votchallenge.net/howto/perfeval.html}} provided by
the organizers of the challenge to compute the scores and the ranking, allowing
us to compare with 38 methods. The interested reader can find more details of
the evaluation protocol in~\cite{Kristan15}.

\vspace{-0.3cm}
\paragraph{OTB dataset.} The online tracking benchmark dataset~\cite{WuY13} is a collection
of 50 of the most commonly used tracking sequences, which are fully annotated. It
contains sequences where the object varies in scale, has fast motion, or is
occluded. The tracking performance on this benchmark is evaluated with a
precision score and area under the curve (AUC) of a success plot, as defined
in~\cite{WuY13}. Precision score is measured as the percentage of frames whose
predicted object location (center of the predicted box) is within a distance of
20 pixels from the center of the ground truth box. Precision plots are also
computed by varying the distance threshold between 0 and 50. Success plots show
the percentage of frames whose intersection over union overlap with the ground
truth annotation is over a threshold varying between 0 and 1. We used the
toolkit\footnote{https://sites.google.com/site/trackerbenchmark/benchmarks/v10}
provided by Wu~\etal~\cite{WuY13} to generate the plots and compute the scores.

\begin{table}
\begin{center}
\begin{tabular}{rlrrr}

\hline
No. &	Method		& Acc.	& Robust. & Avg.\\	
\hline
1  	&Our-ms-rot		&\br{6.07}	&\gr{8.58}	&\rr{7.33}		\\		
2  	&Our-ms			&\rr{4.73}	&10.13		&\br{7.43}		\\	
3  	&DSST~\cite{Danelljan14}			&6.78		&13.99		&\gr{10.39}		\\	
4  	&SAMF			&6.46		&15.65		&11.06			\\
5  	&DGT			&12.67		&10.13		&11.4			\\
6  	&KCF~\cite{Henriques15}			&\gr{6.16}	&16.71		&11.44			\\
7  	&PLT$_{14}$			&16.04		&\br{6.98}	&11.51			\\
8  	&PLT$_{13}$			&19.74		&\rr{4.00}		&11.87			\\
9  	&eASMS			&15.37		&15.1		&15.24			\\
10 	&Our-ss			&16.11		&14.47		&15.29 			\\
11 	&ACAT			&15.1		&16.78		&15.94 			\\
12 	&MatFlow		&23.82		&9.67		&16.74	 		\\	
13 	&HMMTxD			&11.08		&22.51		&16.8 			\\
14 	&MCT			&18.47		&15.14		&16.81 			\\
15 	&qwsEDFT		&19.06		&21.15		&20.11	 		\\	
16 	&ACT			&22.68		&18.1		&20.39 			\\
17 	&ABS			&22.34		&20.49		&21.42 			\\
18 	&VTDMG			&23.22		&19.94		&21.58 			\\
19 	&LGTv1			&31.05		&12.68		&21.87 			\\
20 	&BDF			&24.92		&19.39		&22.15 			\\

\hline

\end{tabular}
\end{center}
\vspace{-0.1cm}
\caption{Performance of the top 20 methods on the VOT2014 dataset. We show the
ranking on accuracy (Acc.) and robustness (Robust.) measures, as well as the
overall rank (Avg.). The top performer in each measure is shown in red, and the
second and third best are in blue and green respectively. We show three
variants of our method -- Our-ms-rot: uses multiscale detector and geometry proposals, Our-ms: uses only multiscale detector proposals, and Our-ss: uses only single-scale detector proposals. The
performance of other methods from the VOT2014 challenge is shown on the project website~\cite{project}.\vspace{-0.5cm}}
\label{tab:vot}
\end{table}

\begin{figure}[t]
\begin{center}
\includegraphics[width=0.8\columnwidth]{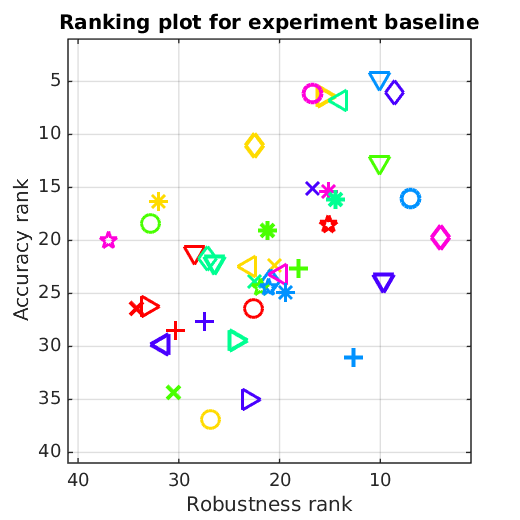}
\includegraphics[width=1.0\columnwidth]{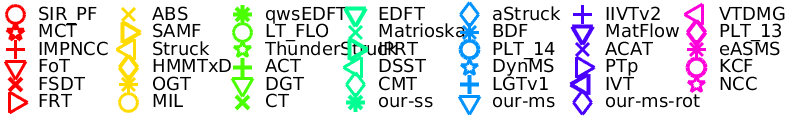}
\end{center}
\vspace{-0.3cm}
\caption{AR ranking plot of all the trackers from the VOT2014 challenge as well
as our approaches. Trackers close to the top right corner of the plot are among
the top performers. Our multiscale approaches (``Our-ms-rot'' and ``Our-ms'')
are closest to the top-right corner compared to all the other
methods.\vspace{-0.4cm}}
\label{fig:arplot}
\end{figure}

\begin{figure*}[t]
\begin{center}
\threefigures{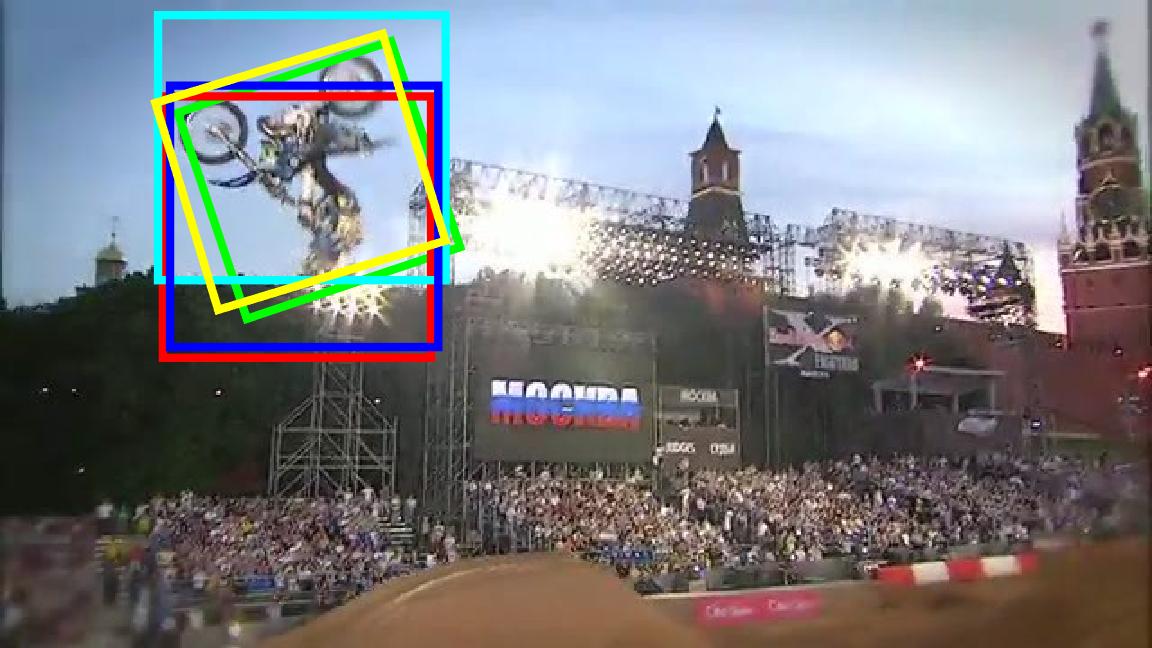}{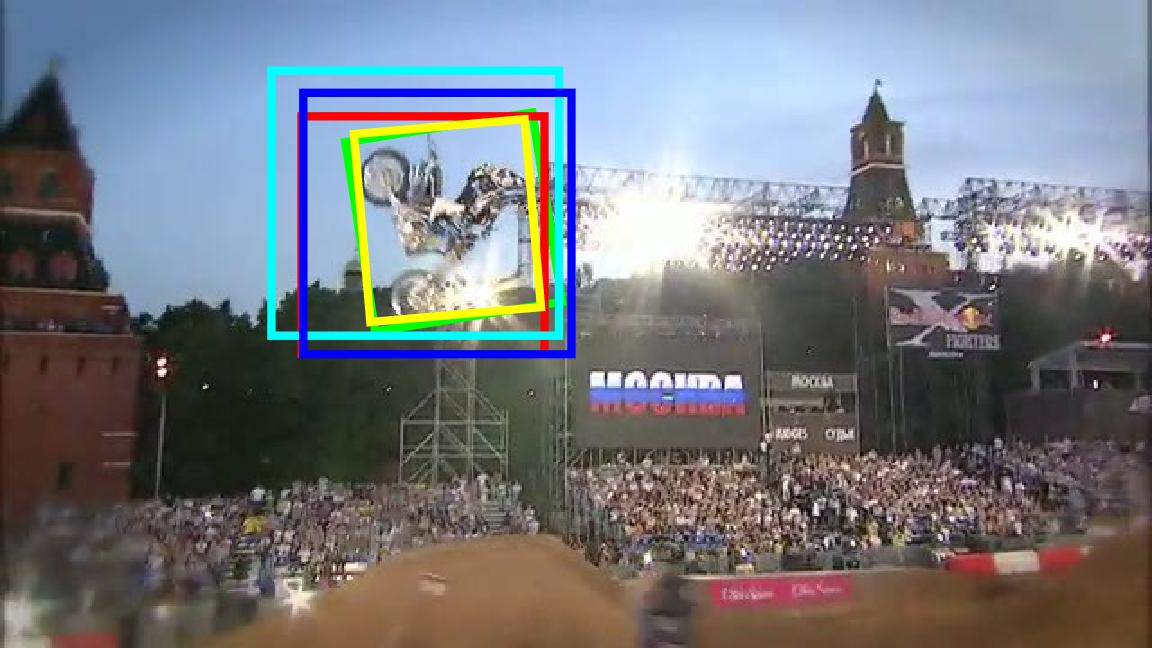}{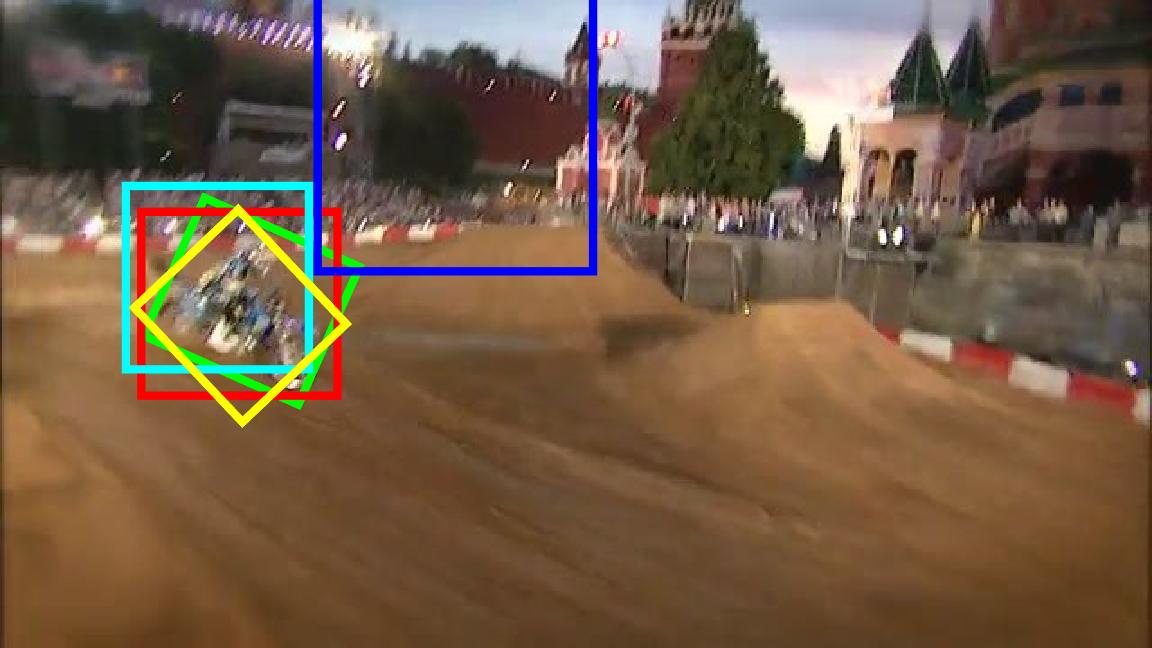}{0.66\columnwidth}
\end{center}
\vspace{-0.7cm}
\caption{Sample result on the {\it motocross} sequence. We compare with
several state-of-the-art trackers.  Ground truth: green, Our result: yellow,
DSST~\cite{Danelljan14}: red, PLT$_{14}$: cyan, Struck~\cite{Hare11}: blue. \it{Best viewed in pdf.} \vspace{-0.4cm}} 
\label{fig:results}
\end{figure*}

\subsection{Results}
\label{sec:results}
\vspace{-0.1cm}
\paragraph{VOT.} The results in Table~\ref{tab:vot} correspond to the baseline
experiment in the toolkit, where each tracker is executed 15 times with the
same ground truth annotation in the first frame, and the overall rank is
computed as described in Section~\ref{sec:dataset}. Overall, our proposal
selection method using detector and geometry proposals (``Our-ms-rot'' in the
table) performs significantly better than DSST~\cite{Danelljan14}, the winner
of the VOT2014 challenge, with an average rank of 7.33 vs 10.39. We also
evaluated two variants of our approach. The first one ``Our-ms'' uses only
the multiscale detector proposals and the second variant ``Our-ss''
is limited to proposals from the detector evaluated only at a single scale. The
overall rank of these two variants is lower than our full method: 7.43 and
15.29 respectively. The variant based on multiscale detector proposals performs
better on the accuracy measure compared to the full method, but is
significantly worse on the robustness measure. In other words, this variant
fails on many more frames than the full method, and benefits from the resulting
reinitializations. Due to significant changes in scale that occur in several
frames in the dataset, the variant based on a single-scale detector performs
rather poorly. CMT~\cite{Nebehay14}, the key point based tracker which
estimates rotation and scale of the object, has an average rank of 24.43 and
is ranked 28th on the list (see~\cite{project}). With the evaluation
protocol of reinitializing the tracker when it fails, we found that using
edgeness measure did not show a significant difference compared to the
detection score. Thus, to keep the comparison with respect to the large number
of trackers (38) reasonable, we show a subset of variants of our methods on
VOT.

On the accuracy measure, our approaches (``Our-ms-rot'' and ``Our-ms'')
are ranked first and second. KCF~\cite{Henriques15} is third, with SAMF and
DSST~\cite{Danelljan14} also performing well. The AR ranking plot in
Figure~\ref{fig:arplot} shows the similar performance of KCF, SAMF and DSST,
where they form a cluster. KCF is a correlation filter based tracker. It is
essentially a regression method trained with HOG features computed on several
patches densely sampled around the object. This method, however, is
significantly inferior on robustness measure (16.71 compared to our result
8.58) as it cannot handle object deformations other than translation. SAMF and
DSST are extensions of the KCF approach. SAMF proposes an adaptive scale for
the patches and also integrates HOG and colour attribute features. It is more
robust than KCF, but still poorer than our result (15.65 vs 8.58).
DSST~\cite{Danelljan14}, the VOT2014 challenge winner, improves the
correlation filtering scheme by estimating the scale of the target and
combining intensity and HOG features. It trains two separate filters: one for
estimating translation and another for scale. While DSST shows better
performance than both SAMF and KCF, it also cannot handle cases where the
target is rotating. Our method addresses this issues and significantly improves
over DSST (8.58 vs 13.99 in robustness rank).

Our tracker is ranked third on the robustness measure. PLT$_{14}$ and
PLT$_{13}$, which are two variants of {\it Struck}~\cite{Hare11}, are first and
second respectively. PLT$_{13}$, the winner of the VOT2013 challenge, uses
object-background colour histograms to weight the features within the bounding
box during SVM training. In essence, it attempts to refine the object
representation from an entire bounding box to an object segmentation.
PLT$_{14}$ is the multiscale version of PLT$_{13}$, and it shows better
accuracy at the cost of a lower robustness. The segmentation cue used in these
methods makes them accurate because it provides a more precise object
representation than a bounding box. However, in many situations, e.g., fast
motion, lack of strong object-background colour priors, it is likely to fail.
This is evident from the significantly lower accuracy of the PLT methods --
16.04, 19.74 compared to 6.07 of our approach (see Table~\ref{tab:vot}).

\begin{figure*}[t]
\begin{center}
\twofigures{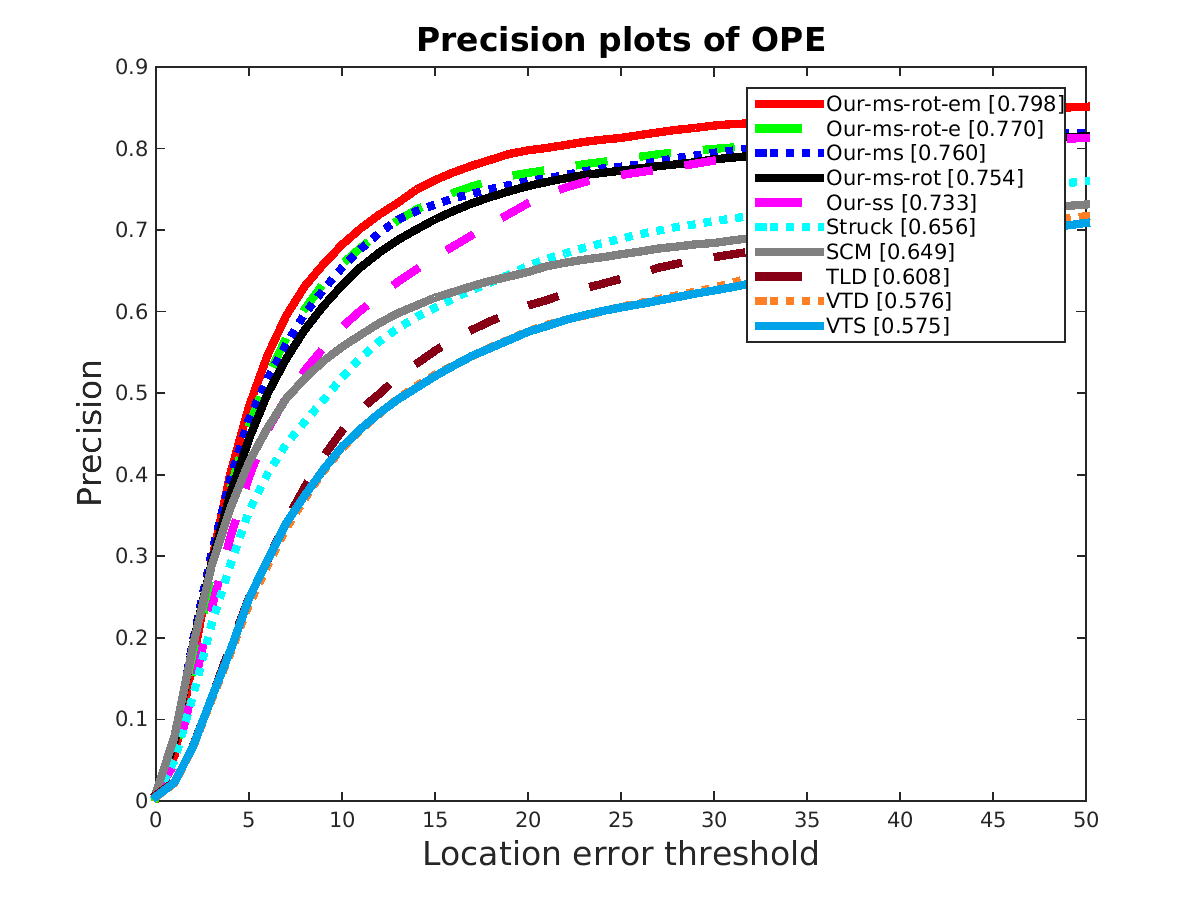}{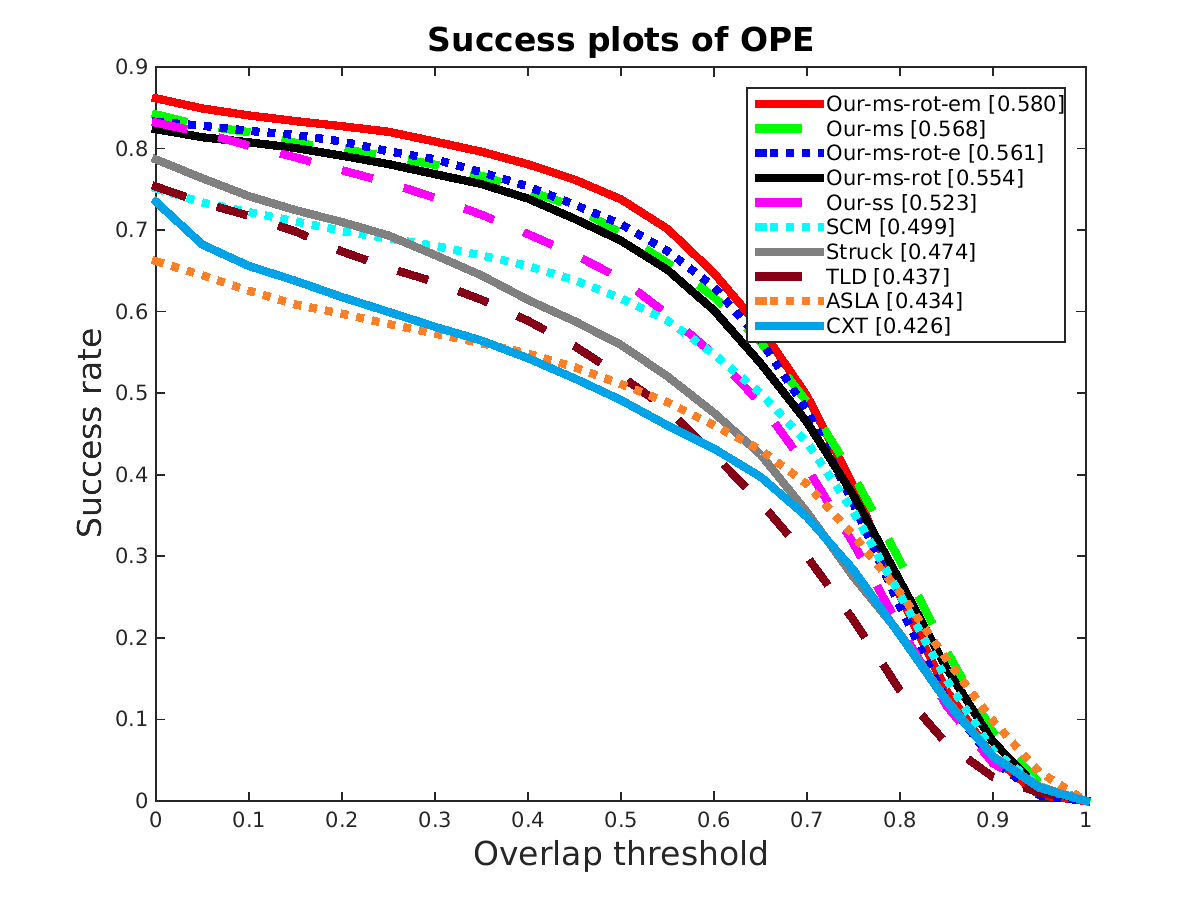}{1.1\columnwidth}
\end{center}\vspace{-1cm}
\caption{Precision (left) and success rate (right) plots on the OTB dataset.
For clarity we compare to the top 10 algorithms in each figure. The complete plots
are on the project website~\cite{project}. The precision plots are summarized 
with the precision score at 20 pixel error threshold and 
the  success plots with the area under the curve.
These values are shown in the legend. \it{Best
viewed in pdf.}\vspace{-0.5cm}}
\label{fig:otbplots}
\end{figure*}

We present a sample qualitative result in
Figure~\ref{fig:results}, see~\cite{project} for additional results. 
The {\it motocross}
sequence is considered as one of the most challenging sequences in this
benchmark, based on the number of trackers that fail on it. Our approach
performs significantly better other methods. We compare to the results of DSST and
PLT$_{14}$, which performed well on the VOT2014 challenge, and also {\it Struck},
the top performer on several datasets.


\vspace{-0.35cm}
\paragraph{OTB.}
The results in Figure~\ref{fig:otbplots} correspond to the one-pass evaluation
(OPE) of the toolkit~\cite{WuY13}. The trackers are tested on more than 29,000
frames in this evaluation. We show the success and precision plots of our
method in Figure~\ref{fig:otbplots} and compare it with the top 10 methods.
Comparison with all the 28 trackers evaluated is shown in~\cite{project}. Our overall method ``Our-ms-rot-em'' and its four variants outperform
all the trackers on both the measures. More precisely, our tracker achieves
0.798 precision score compared to 0.656 of {\it Struck}~\cite{Hare11}, and
0.580 AUC of success plot compared to 0.499 of SCM~\cite{Zhong12} -- the top
two performers on this benchmark dataset~\cite{WuY13}.

Our single-scale tracker ``Our-ss'' is inferior to the other four variants:
0.733 precision score and 0.523 AUC of success plot due to significant changes
in scale and/or rotation in several sequences. The performance of our multiscale version
without geometry proposals ``Our-ms'', and the multiscale one with geometry
proposals ``Our-ms-rot'' is very similar. This is expected, despite
``Our-ms-rot'' performing better than ``Our-ms'' on the VOT dataset, because
the ground truth annotations in the OTB dataset are axis-aligned and not
rotated boxes, as is the case in VOT. Thus, a rotating object can be enclosed
(although imprecisely) by a larger axis-aligned box. This makes ``Our-ms'' as
effective as ``Our-ms-rot'' following the OTB evaluation protocol. 
In Figure~\ref{fig:otbplots} we also show the
influence of scores computed with edges ``e'' and motion boundaries ``m''. The
precision result using (multiscale) detector confidence alone to select from
the candidate set (``Our-ms-rot'') is 0.754. Incorporating the measure computed
with edges (``Ours-ms-rot-e'') improves this by 2.1\% to 0.770, and the cue
computed with motion boundary (``Our-ms-rot-em'') gives a further 3.6\%
improvement to 0.798. A similar improvement pattern can be seen on the success
plot.

A couple of recent papers~\cite{Hua14,Supancic13} also evaluated their trackers
on the OTB dataset. However, they are not online trackers, in that they use the
information for subsequent frames to infer the object location in the current
frame. We, nevertheless, compare to them using mean $F_1$ score
(with 50\% overlap threshold), following the protocol in~\cite{Hua14}. Our
tracker (``Our-ms-rot-em'') performs significantly better with a score of
0.757 compared to 0.657 and 0.661 of~\cite{Hua14} and~\cite{Supancic13}
respectively.

\begin{figure}[t]
\begin{center}
\includegraphics[width=0.8\columnwidth]{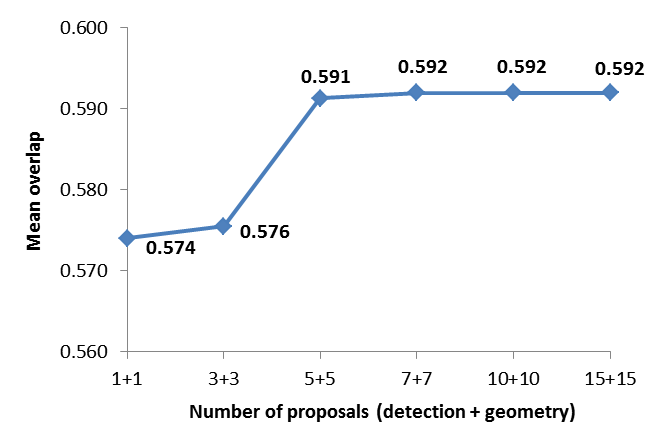}
\end{center}\vspace{-0.3cm}
\caption{Influence of the number of (detection and geometry) proposals on
the tracking performance on the OTB dataset.\vspace{-0.5cm}}
\label{fig:propvsperf}
\end{figure}

In Figure~\ref{fig:propvsperf} we evaluate the performance of our approach
(``Our-ms-rot-em'') with respect to the number of candidate proposals on the
OTB dataset. Here, we measure the performance as mean overlap between the
predicted and the ground truth bounding boxes over the entire sequence. We observe
that the mean overlap score increases initially with the number of proposals,
but then quickly saturates. In other words, we require a minimum number of
proposals (5 each for detection and geometry, which we use in this paper) for
tracking an object effectively. Adding further candidate proposals beyond this
does not affect the performance, as the new candidates are either very similar
to or less accurate than those already in the proposal set.

\vspace{-0.3cm}
\paragraph{Computation time.}
Our Matlab and mex implementation is un-optimized and not real-time. It
performs as follows (average fps on VOT+OTB datasets): our-ss:~6.4,
our-ms:~4.1, our-ms-rot:~3.1, our-ms-rot-e:~2.1, our-ms-rot-em:~1.9 on a
cluster node with 20 cores. The last three variants require pre-computed
optical flow~\cite{Brox11}, which runs at 0.3 fps on a GPU. Our implementation
can certainly be sped up in several ways, e.g., by scaling the images down,
speeding up the feature extraction process.


\vspace{-0.2cm}
\section{Summary}
\label{sec:summary}
\vspace{-0.2cm}
In this paper, we present a new tracking-by-detection framework for online
object tracking. Our approach begins with building a candidate set of object
location proposals extracted using a learned detector model. It is then
augmented with novel proposals computed by estimating the geometric
transformation(s) undergone by the object. We localize the object by selecting
the best proposal from this candidate set using multiple cues: detection
confidence score, edges and motion boundaries. The performance of our tracker is
evaluated extensively on the VOT2014 challenge and the OTB datasets. We show
state-of-the-art results on both these benchmarks, significantly improving over
the top performers of these two evaluations.

\fboxrule=2pt

\vspace{-0.3cm}
\paragraph{Acknowledgements.}
This work was supported in part by the MSR-Inria joint project, Google Faculty
Research Award, and the ERC advanced grant ALLEGRO.
\vspace{-0.2cm}

\renewcommand\refname{References\vskip -0.2cm}
{\small
\bibliographystyle{ieee}
\bibliography{paper}
}

\end{document}